\PassOptionsToPackage{unicode}{hyperref}
\PassOptionsToPackage{hyphens}{url}
\documentclass[
]{article}
\usepackage{amsmath,amssymb}
\usepackage{lmodern}
\usepackage{iftex}
\ifPDFTeX
  \usepackage[T1]{fontenc}
  \usepackage[utf8]{inputenc}
  \usepackage{textcomp} 
\else 
  \usepackage{unicode-math}
  \defaultfontfeatures{Scale=MatchLowercase}
  \defaultfontfeatures[\rmfamily]{Ligatures=TeX,Scale=1}
\fi
\IfFileExists{upquote.sty}{\usepackage{upquote}}{}
\IfFileExists{microtype.sty}{
  \usepackage[]{microtype}
  \UseMicrotypeSet[protrusion]{basicmath} 
}{}
\makeatletter
\@ifundefined{KOMAClassName}{
  \IfFileExists{parskip.sty}{%
    \usepackage{parskip}
  }{
    \setlength{\parindent}{0pt}
    \setlength{\parskip}{6pt plus 2pt minus 1pt}}
}{
  \KOMAoptions{parskip=half}}
\makeatother
\usepackage{xcolor}
\IfFileExists{xurl.sty}{\usepackage{xurl}}{} 
\IfFileExists{bookmark.sty}{\usepackage{bookmark}}{\usepackage{hyperref}}
\hypersetup{
  pdftitle={Sentiment Analysis with R},
  pdfauthor={a Working Paper and Tutorial by Dennis Klinkhammer},
  hidelinks,
  pdfcreator={LaTeX via pandoc}}
\usepackage[margin=1in]{geometry}
\usepackage{color}
\usepackage{fancyvrb}

\DefineVerbatimEnvironment{Highlighting}{Verbatim}{commandchars=\\\{\}}
\usepackage{framed}
\definecolor{shadecolor}{RGB}{248,248,248}
\newenvironment{Shaded}{\begin{snugshade}}{\end{snugshade}}

\newcommand{\AttributeTok}[1]{\textcolor[rgb]{0.77,0.63,0.00}{#1}}

\newcommand{\CommentTok}[1]{\textcolor[rgb]{0.56,0.35,0.01}{\textit{#1}}}

\newcommand{\ConstantTok}[1]{\textcolor[rgb]{0.00,0.00,0.00}{#1}}

\newcommand{\DecValTok}[1]{\textcolor[rgb]{0.00,0.00,0.81}{#1}}

\newcommand{\FloatTok}[1]{\textcolor[rgb]{0.00,0.00,0.81}{#1}}
\newcommand{\FunctionTok}[1]{\textcolor[rgb]{0.00,0.00,0.00}{#1}}

\newcommand{\NormalTok}[1]{#1}

\newcommand{\OtherTok}[1]{\textcolor[rgb]{0.56,0.35,0.01}{#1}}

\newcommand{\SpecialCharTok}[1]{\textcolor[rgb]{0.00,0.00,0.00}{#1}}

\newcommand{\StringTok}[1]{\textcolor[rgb]{0.31,0.60,0.02}{#1}}

\usepackage{graphicx}
\makeatletter
\def\maxwidth{\ifdim\Gin@nat@width>\linewidth\linewidth\else\Gin@nat@width\fi}
\def\maxheight{\ifdim\Gin@nat@height>\textheight\textheight\else\Gin@nat@height\fi}
\makeatother
\setkeys{Gin}{width=\maxwidth,height=\maxheight,keepaspectratio}
\makeatletter
\def\fps@figure{htbp}
\makeatother
\ifLuaTeX
  \usepackage{selnolig}  
\fi

\title{Sentiment Analysis with R}
\usepackage{etoolbox}
\makeatletter
\providecommand{\subtitle}[1]{
  \apptocmd{\@title}{\par {\large #1 \par}}{}{}
}
\makeatother
\subtitle{Natural Language Processing for Semi-Automated Assessments of
Qualitative Data}
\author{a Working Paper and Tutorial by Dennis Klinkhammer}
\date{}

\begin{document}
\maketitle
\begin{abstract}
Sentiment analysis is a sub-discipline in the field of natural language
processing and computational linguistics and can be used for automated
or semi-automated analyses of text documents. One of the aims of these
analyses is to recognize an expressed attitude as positive or negative
as it can be contained in comments on social media platforms or
political documents and speeches as well as fictional and nonfictional
texts. Regarding analyses of comments on social media platforms, this is
an extension of the previous tutorial on semi-automated screenings of
social media network data. A longitudinal perspective regarding social
media comments as well as cross-sectional perspectives regarding
fictional and nonfictional texts, e.g.~entire books and libraries, can
lead to extensive text documents. Their analyses can be simplified and
accelerated by using sentiment analysis with acceptable inter-rater
reliability. Therefore, this tutorial introduces the basic functions for
performing a sentiment analysis with R and explains how text documents
can be analysed step by step - regardless of their underlying
formatting. All prerequisites and steps are described in detail and
associated codes are available on GitHub. A comparison of two political
speeches illustrates a possible use case.
\end{abstract}

\hypertarget{keywords}{%
\paragraph{Keywords}\label{keywords}}

Sentiment Analysis, Natural Language Processing, Computational
Linguistics, Qualitative Data

\hypertarget{introduction-to-sentiment-analysis-and-this-tutorial}{%
\subsection{Introduction to Sentiment Analysis and this
Tutorial}\label{introduction-to-sentiment-analysis-and-this-tutorial}}

Sentiment analysis -- as the name suggests - can be used to capture the
sentiment in qualitative data, such as text documents. Text documents
can contain different types of content and information, e.g.~comments on
social media platforms or political documents and speeches as well as
fictional and nonfictional texts up to entire libraries. Usually, these
text documents come in different formats, e.g.~PDF format, HTML format
and many more, depending on the medium the text is located. Since most
formats can be converted into a simpler format, such as the TXT format,
this tutorial uses the TXT format as default. For example, if someone
wants to convert a text document from PDF format to TXT format, the copy
and paste function would be sufficient for this tutorial.

Within this TXT format the polarity can be classified word by word as
positive or negative and in some cases neutral via basic sentiment
analysis. In addition, different types of emotional states can be
classified via advanced sentiment analysis and by using the NRC
Word-Emotion Association Lexicon, the first word-emotion lexicon with
eight basic emotional states (Mohammad 2020). Despite positive and
negative sentiments, anger, fear, anticipation, trust, surprise,
sadness, joy, and disgust can be classified as well. Furthermore, a
semi-automated sentiment analysis has a sufficient
inter-rater-reliability with less time requirements. The
inter-rater-reliability is a degree of agreement among independent
observers who rate, code, or assess the same phenomenon within a text
document. While scientist usually achieve an inter-rater-reliability up
to 80\% the semi-automated ones can achieve up to 70\%. This appears to
be an acceptable value, because even if different types of
semi-automated sentiment analyses would agree up to 100\%, research
indicates that scientists would still disagree by 20\% (Ogneva 2010).
Therefore, sentiment analyses can be found in a broad application
context, of which a few will be presented in this tutorial.

Since this is an addition to the previous tutorial on semi-automated
screenings of social media network data (Klinkhammer 2020), the analysis
of comments on social media platforms should not go unmentioned.
Especially social media platforms offer several users a low-threshold
opportunity to exchange opinions and experiences. For example, these
opinions and experiences can affect various areas of society, such as
political and economical ones. From a methodological point of view, the
question whether one technique is equally suitable for entire books as
well as short comments on social media platforms seems relevant.
Research indicates that, for example, comments on social media platforms
can be used to capture social issues like radicalisation and extemism
(Tanoli et al.~2022), sexuality (Wood et al.~2017), side effects of
medication and drugs (Korkontzelos et al.~2016) as well as for the
reflection of the offline political landscape (Tumasjan et al.~2010). A
well known use-case is the political campaign of former U.S. president
Barack Obama, who used sentiment analysis back in 2012. These
methodological approaches hardly differ from the analyses of entire
books. There are many possible use-cases, but also numerous challenges
in the application of sentiment analysis. Accordingly, research in this
area continues (Hamborg \& Donnay 2021) and this tutorial refers to a
semi-automated sentiment analysis rather than automated ones.

This contribution includes all software requirements, a full disclosure
of codes for the R programming language, the entire process of accessing
and pre-precessing text documents and how to perform basic and advances
sentiment analyses. This tutorial is mainly supposed to be a
methodological tutorial for students, and researchers.

\hypertarget{software-requirements}{%
\subsection{Software Requirements}\label{software-requirements}}

Written in \emph{R Markdown}, this tutorial refers to the R programming
language. A free software environment for using R is available for
Linux, macOS by Apple and Windows by Microsoft. The main purpose of R is
statistical computing and it is both used for manual quantitative and
qualitative analyses as well as automated or semi-automated analyses.
When it comes to Big Data, R can also be used for unsupervised and
supervised Machine Learning.

In detail, R is an object based programming language. Therefore,
datasets, variables, cases, values as well as functions can be applied
as a combination of objects. All commands, as combinations of functions,
datasets, variables, cases, values and functions will be highlighted, so
that they can be used as step by step tutorial. The commands have to be
entered directly into the R terminal, which is available after
downloading, installing and starting the R software environment.

\hypertarget{preparations-attaching-necessary-packages}{%
\subsection{Preparations: Attaching necessary
Packages}\label{preparations-attaching-necessary-packages}}

This tutorial requires six additional packages in order to expand the
range of basic R functions. All packages can be installed by using the
\emph{install(\ldots)} command and attached via the
\emph{library(\ldots)} command by typing the following commands directly
into the R terminal.

Since the analysis of extensive text documents requires a focus on every
single element that is to be analysed, it is necessary to break down the
underlying data structure into manageable little pieces. A package that
is specifically designed to do so is called \emph{dplyr}. It can split,
apply and combine data for further analytical steps (Wickham 2022). The
package \emph{dplyr} can be installed and attached as follows:

\begin{Shaded}
\begin{Highlighting}[]
\FunctionTok{install.packages}\NormalTok{(}\StringTok{"dplyr"}\NormalTok{, }\AttributeTok{dependencies=}\ConstantTok{TRUE}\NormalTok{)}
\FunctionTok{library}\NormalTok{(dplyr)}
\end{Highlighting}
\end{Shaded}

The second package is called \emph{stringr}. Since qualitative data,
like the text in social media comments, is represented by character
variables in R, a package that can process and - if necessary -
manipulate individual characters within the strings of a character
variable is required (Wickham 2019); A string is marked either by single
quote signs or double quote signs. In order to install and attach the
\emph{stringr} package, following commands can be typed in the R
terminal:

\begin{Shaded}
\begin{Highlighting}[]
\FunctionTok{install.packages}\NormalTok{(}\StringTok{"stringr"}\NormalTok{, }\AttributeTok{dependencies=}\ConstantTok{TRUE}\NormalTok{)}
\FunctionTok{library}\NormalTok{(stringr)}
\end{Highlighting}
\end{Shaded}

Another necessary package is called \emph{textdata}. It contains several
words as references and sentiment libraries, such as the NRC
Word-Emotion Association Lexicon. So it does not only needed to be
installed by the \emph{install.packages(\ldots)} command and attached by
the subsequent \emph{library(\ldots)} command, but the NRC Word-Emotion
Association Lexicon must be installed for advanced sentiment analysis as
well. In addition, a second lexicon will be installed to implement basic
sentiment analysis: The Bing Sentiment Lexicon. This can be done via the
\emph{get\_sentiments(\ldots)} command:

\begin{Shaded}
\begin{Highlighting}[]
\FunctionTok{install.packages}\NormalTok{(}\StringTok{"textdata"}\NormalTok{, }\AttributeTok{dependencies=}\ConstantTok{TRUE}\NormalTok{)}
\FunctionTok{library}\NormalTok{(textdata)}
\FunctionTok{get\_sentiments}\NormalTok{(}\StringTok{"nrc"}\NormalTok{)}
\FunctionTok{get\_sentiment}\NormalTok{(}\StringTok{"bing"}\NormalTok{)}
\end{Highlighting}
\end{Shaded}

Sentiment analysis is a text mining technique and the package
\emph{tidytext} is required in order to convert conventional text
documents into tidy formats, such as single words without punctuation or
spaces (De Queiroz et al.~2022). This allows scientists to focus on
paragraphs or otherwise separated content word by word. As a result, the
tidy text format lists and counts all words individually and assigns
them a numbered line according to their paragraph or other used methods
of content separation. Again, this package can be installed and attached
as follows:

\begin{Shaded}
\begin{Highlighting}[]
\FunctionTok{install.packages}\NormalTok{(}\StringTok{"tidytext"}\NormalTok{, }\AttributeTok{dependencies=}\ConstantTok{TRUE}\NormalTok{)}
\FunctionTok{library}\NormalTok{(tidytext)}
\end{Highlighting}
\end{Shaded}

The package \emph{tidytext} provides the connection between the packages
\emph{dplyr} and \emph{ggplot2} by using their basic formulas and
commands. The latter is responsible for the detailed visualisation of
sentiments and other types of results, based on ``The Grammer of
Graphics'' (Wickham et al.~2022). In particular, defining the details of
a visualisation enables scientists to create informative as well as
attractive plots. Therefore, the package \emph{ggplot2} requires several
dependencies in order to carry out this task:

\begin{Shaded}
\begin{Highlighting}[]
\FunctionTok{install.packages}\NormalTok{(}\StringTok{"ggplot2"}\NormalTok{, }\AttributeTok{dependencies=}\ConstantTok{TRUE}\NormalTok{)}
\FunctionTok{library}\NormalTok{(ggplot2)}
\end{Highlighting}
\end{Shaded}

Finally, the package \emph{gridExtra} enables scientists to arrange
multiple visualisations at once and to create dashboards for an
intuitive display of relevant information (Auguie \& Antonov 2017).
Following commands must be typed in the R terminal in order to make the
package \emph{gridExtra} work:

\begin{Shaded}
\begin{Highlighting}[]
\FunctionTok{install.packages}\NormalTok{(}\StringTok{"gridExtra"}\NormalTok{, }\AttributeTok{dependencies=}\ConstantTok{TRUE}\NormalTok{)}
\FunctionTok{library}\NormalTok{(gridExtra)}
\end{Highlighting}
\end{Shaded}

Additional note: If not already pre-installed, the package
\emph{magrittr} is also required in order to make use of the
forward-pipe operator for more elegant coding. It is possible that
further packages have to be installed in the R environment. This will be
automatically checked via the extension of the \emph{install (\ldots)}
command with \emph{dependencies = TRUE} that will install additional
packages, if necessary. In a freshly created R environment (based upon
version 4.1.2 of R), the six packages listed above have been running
sufficient on Ubuntu 22.04 LTS (Linux Kernel 5.15.0-37), macOS Monterey
(12.4) and Windows 11 (22000.708), each brought to application on
RStudio.

\hypertarget{data-pre-processing-getting-and-cleaning-text-documents}{%
\subsection{Data Pre-Processing: Getting and Cleaning Text
Documents}\label{data-pre-processing-getting-and-cleaning-text-documents}}

Data pre-processing is used to check datasets for irrelevant and
redundant information present or noisy and unreliable data. In a first
step, an external text document is imported into the R working
environment by using the \emph{read.delim(\ldots)} command and saving
this text document as a new object called \emph{imported\_text}. The
external text document is TXT formatted and can be accessed by entering
the associated directory and file name. In this case, a chapter from one
of the Sherlock Holmes novels will be imported (thanks to the great work
of Arthur Conan Doyle):

\begin{Shaded}
\begin{Highlighting}[]
\NormalTok{imported\_text }\OtherTok{\textless{}{-}} \FunctionTok{read.delim}\NormalTok{(}\StringTok{"your\_unformatted\_document.txt"}\NormalTok{, }\AttributeTok{header=}\NormalTok{F, }\AttributeTok{sep=}\StringTok{"}\SpecialCharTok{\textbackslash{}t}\StringTok{"}\NormalTok{)}
\end{Highlighting}
\end{Shaded}

The subsequent \emph{dim(\ldots)} command indicates the number of
paragraphs within this text document, which is necessary for the next
step, as well as the number of variables within the
\emph{imported\_text}. As a result, \emph{229} paragraphs and one
Variable, named \emph{V1}, can be processed further:

\begin{Shaded}
\begin{Highlighting}[]
\FunctionTok{dim}\NormalTok{(imported\_text)}
\end{Highlighting}
\end{Shaded}

\begin{verbatim}
## [1] 229   1
\end{verbatim}

The next step creates tibbles out of the \emph{imported\_text}. Tibbles
are data frames that are considered plain and simple. In fact, they are
so plain and simple, that the number of lines needs to be entered
manually in order to define the structure of the data frame. In this
case, the number of lines within the tibble is supposed to be the exact
number of paragraphs of the \emph{imported\_text}, hence \emph{229}. As
a result, the \emph{tibble(\ldots)} command leads to a new object called
\emph{text\_df} and the first six lines can be inspected by using the
\emph{head(\ldots)} command:

\begin{Shaded}
\begin{Highlighting}[]
\NormalTok{text\_df }\OtherTok{\textless{}{-}} \FunctionTok{tibble}\NormalTok{(}\AttributeTok{line=}\DecValTok{1}\SpecialCharTok{:}\DecValTok{229}\NormalTok{, }\AttributeTok{text=}\NormalTok{imported\_text}\SpecialCharTok{$}\NormalTok{V1)}
\FunctionTok{head}\NormalTok{(text\_df)}
\end{Highlighting}
\end{Shaded}

\begin{verbatim}
## # A tibble: 6 x 2
##    line text                                                                   
##   <int> <chr>                                                                  
## 1     1 "To Sherlock Holmes she is always the woman. I have seldom heard him"  
## 2     2 "     mention her under any other name. In his eyes she eclipses and"  
## 3     3 "     predominates the whole of her sex. It was not that he felt any"  
## 4     4 "     emotion akin to love for Irene Adler. All emotions, and that one"
## 5     5 "     particularly, were abhorrent to his cold, precise but admirably" 
## 6     6 "     balanced mind. He was, I take it, the most perfect reasoning and"
\end{verbatim}

Although the object \emph{text\_df} is plain and simple formatted, it
needs to be pre-processed further. Sentiment analysis requires a tidy
data set, which can be generated by using the object \emph{text\_df}
combined with the \emph{unnest.tokens(\ldots)} command. In this case,
each word within the text has to be separated, indicated by \emph{word}
and \emph{text}. For the first time in this tutorial the connection
between an object and command will be established via the forward-pipe
operator \emph{\%\textgreater\%}. Later on, this technique allows to
combine several commands regarding one object and - in this case - to
generate the new object \emph{text\_tidy}:

\begin{Shaded}
\begin{Highlighting}[]
\NormalTok{text\_tidy }\OtherTok{\textless{}{-}}\NormalTok{ text\_df }\SpecialCharTok{\%\textgreater{}\%}
  \FunctionTok{unnest\_tokens}\NormalTok{(word, text)}
\end{Highlighting}
\end{Shaded}

This object is just a listing of all the words in the different
paragraphs. This allows the sentiment analysis to be performed word by
word and line by line. Since non-essential words may be included in this
list, they can be eliminated by using the \emph{stop\_words} data via
the \emph{data(\ldots)} command. Again, the forward-pipe operator
combines the object \emph{text\_tidy} with the new command
\emph{anti\_join(\ldots)} in order to exclude the specified words:

\begin{Shaded}
\begin{Highlighting}[]
\FunctionTok{data}\NormalTok{(stop\_words)}
\NormalTok{text\_tidy }\OtherTok{\textless{}{-}}\NormalTok{ text\_tidy }\SpecialCharTok{\%\textgreater{}\%}
  \FunctionTok{anti\_join}\NormalTok{(stop\_words)}
\end{Highlighting}
\end{Shaded}

\begin{verbatim}
## Joining, by = "word"
\end{verbatim}

Now the object \emph{text\_tidy} is prepared for sentiment analysis. In
addition, custom words can also be excluded from this object. They can
be generated by using the \emph{tibble(\ldots)} command and repeating
the previous step of the pre-processing.

\begin{Shaded}
\begin{Highlighting}[]
\NormalTok{custom\_stop\_words }\OtherTok{\textless{}{-}} \FunctionTok{c}\NormalTok{(}\StringTok{"custom\_word"}\NormalTok{)}
\NormalTok{custom\_stop\_words }\OtherTok{\textless{}{-}} \FunctionTok{tibble}\NormalTok{(}\DecValTok{1}\SpecialCharTok{:}\DecValTok{1}\NormalTok{, }\AttributeTok{word=}\NormalTok{custom\_stop\_words)}
\end{Highlighting}
\end{Shaded}

Finally, the \emph{head(\ldots)} command generates an output of the
first six lines of the pre-processed object \emph{text\_tidy}. This
object includes only words that are relevant for an understanding of the
original text document and the lines - respectively paragraphs - they
have been mentioned within the text document:

\newpage

\begin{Shaded}
\begin{Highlighting}[]
\FunctionTok{head}\NormalTok{(text\_tidy)}
\end{Highlighting}
\end{Shaded}

\begin{verbatim}
## # A tibble: 6 x 2
##    line word    
##   <int> <chr>   
## 1     1 sherlock
## 2     1 holmes  
## 3     1 woman   
## 4     1 seldom  
## 5     1 heard   
## 6     2 mention
\end{verbatim}

\hypertarget{analysis---1-identification-of-common-words-and-their-sentiment}{%
\subsection{Analysis - 1: Identification of Common Words and their
Sentiment}\label{analysis---1-identification-of-common-words-and-their-sentiment}}

The first three analytical steps refer to the the NRC Word-Emotion
Association Lexicon, which can be specified as criteria for sentiment
analysis via the \emph{inner\_join(\ldots)} command. Since the first
analytical step of this tutorial relates to the frequency of the words
used within the text document and their sentiments as well as the
expressed emotional states, the \emph{count(\ldots)} command counts and
sorts each \emph{word} and its \emph{sentiment} regarding the underlying
object \emph{text\_tidy}. The result of this step will be saved as a new
object \emph{nrc\_word\_counts}:

\begin{Shaded}
\begin{Highlighting}[]
\CommentTok{\# Positive und negative Wörter auflisten}
\NormalTok{nrc\_word\_counts }\OtherTok{\textless{}{-}}\NormalTok{ text\_tidy }\SpecialCharTok{\%\textgreater{}\%}
  \FunctionTok{inner\_join}\NormalTok{(}\FunctionTok{get\_sentiments}\NormalTok{(}\StringTok{"nrc"}\NormalTok{)) }\SpecialCharTok{\%\textgreater{}\%}
  \FunctionTok{count}\NormalTok{(word, sentiment, }\AttributeTok{sort =} \ConstantTok{TRUE}\NormalTok{) }\SpecialCharTok{\%\textgreater{}\%}
  \FunctionTok{ungroup}\NormalTok{()}
\end{Highlighting}
\end{Shaded}

\begin{verbatim}
## Joining, by = "word"
\end{verbatim}

This new object can be converted into a corresponding graphic. The words
shall be listed in descending order according to their frequency, where
the minimum number is specified as \emph{n \textgreater{} 2} within the
\emph{filter(\ldots)} command. Larger text documents can contain
multiple repetitions of the same words, so that the minimum number has
to be adapted accordingly. It is advisable to try different numbers if
necessary. The order is specified by using the \emph{mutate(\ldots)}
command and the \emph{ggplot(\ldots)} command combines the words to be
highlighted with their sentiment and emotional state, where each is
represented by an individual color, specified within the
\emph{geom\_col(\ldots)} command. Since the sentiments and emotional
states are to be shown as horizontal bar charts, the command
\emph{coord\_flip} is also used. The plot also requires a label that has
to be assigned to the Y-axis. This is done by using the
\emph{labs(\ldots)} command. A title will be added via the
\emph{ggtitle(\ldots)} command:

\begin{Shaded}
\begin{Highlighting}[]
\NormalTok{nrc\_word\_counts }\SpecialCharTok{\%\textgreater{}\%}
  \FunctionTok{filter}\NormalTok{(n }\SpecialCharTok{\textgreater{}} \DecValTok{2}\NormalTok{) }\SpecialCharTok{\%\textgreater{}\%}
  \FunctionTok{mutate}\NormalTok{(}\AttributeTok{word =} \FunctionTok{reorder}\NormalTok{(word, n)) }\SpecialCharTok{\%\textgreater{}\%}
  \FunctionTok{ggplot}\NormalTok{(}\FunctionTok{aes}\NormalTok{(word, n, }\AttributeTok{fill =}\NormalTok{ sentiment)) }\SpecialCharTok{+}
  \FunctionTok{geom\_col}\NormalTok{() }\SpecialCharTok{+}
  \FunctionTok{coord\_flip}\NormalTok{() }\SpecialCharTok{+}
  \FunctionTok{labs}\NormalTok{(}\AttributeTok{y =} \StringTok{"sentiment (n)"}\NormalTok{) }\SpecialCharTok{+}
  \FunctionTok{ggtitle}\NormalTok{(}\StringTok{"Common Words \& Sentiments (Frequency)"}\NormalTok{)}
\end{Highlighting}
\end{Shaded}

\includegraphics{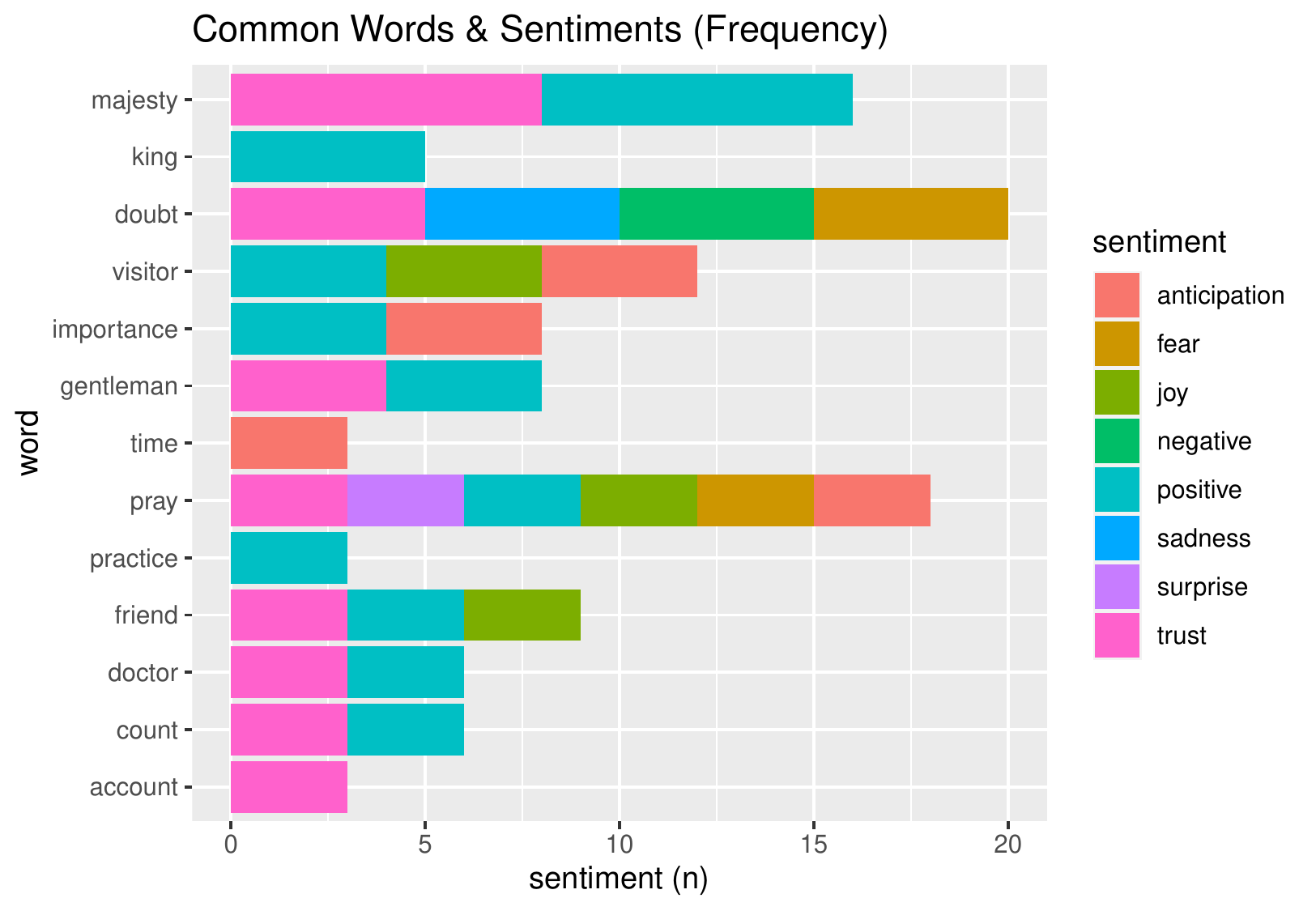}

The result is a graphical representation of the most common words in
descending order (Y-axis) and the frequency of associated sentiments and
emotional states (X-axis). For example, the word \emph{doubt} is the
third most common word but it represents the emotional states trust,
sadness and fear as well as a negative sentiment within different
paragraphs of the text document.

\hypertarget{analysis---2-distribution-of-sentiments}{%
\subsection{Analysis - 2: Distribution of
Sentiments}\label{analysis---2-distribution-of-sentiments}}

In the second step of the analysis the total number of each sentiment
and emotional state within the text document shall be focused in order
to highlight their distribution. Here the object
\emph{nrc\_word\_counts} is used again, which can be directly specified
and plotted via the pipe-forward operator. This time the words will not
be specified within the \emph{ggplot(\ldots)} command, but the
sentiments and emotional states via \emph{sentiment}:

\begin{Shaded}
\begin{Highlighting}[]
\NormalTok{nrc\_word\_counts }\SpecialCharTok{\%\textgreater{}\%}
  \FunctionTok{inner\_join}\NormalTok{(}\FunctionTok{get\_sentiments}\NormalTok{(}\StringTok{"nrc"}\NormalTok{)) }\SpecialCharTok{\%\textgreater{}\%}
  \FunctionTok{count}\NormalTok{(word, sentiment) }\SpecialCharTok{\%\textgreater{}\%}
  \FunctionTok{ggplot}\NormalTok{(}\FunctionTok{aes}\NormalTok{(sentiment, n, }\AttributeTok{fill =}\NormalTok{ sentiment)) }\SpecialCharTok{+}
  \FunctionTok{geom\_col}\NormalTok{(}\AttributeTok{show.legend =} \ConstantTok{FALSE}\NormalTok{) }\SpecialCharTok{+}
  \FunctionTok{ggtitle}\NormalTok{(}\StringTok{"Sentiments (Distribution)"}\NormalTok{) }\SpecialCharTok{+}
  \FunctionTok{coord\_flip}\NormalTok{()}
\end{Highlighting}
\end{Shaded}

\begin{verbatim}
## Joining, by = c("word", "sentiment")
\end{verbatim}

\includegraphics{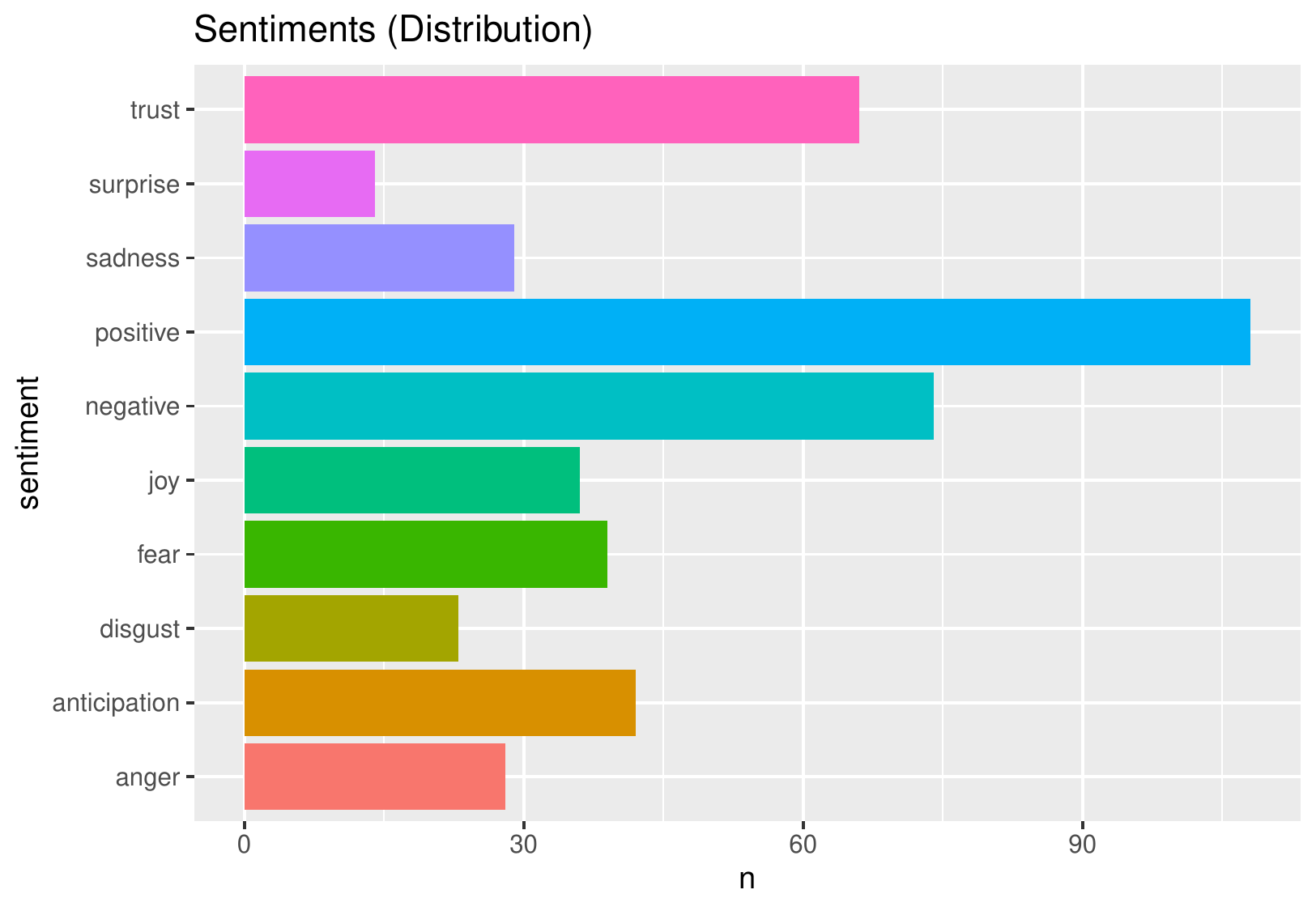}

This output highlights the total number (X-axis) of each sentiment and
emotional state (Y-axix) within the entire chapter of the Sherlock
Holmes novel. Overall, the chapter is dominated by a positive sentiment.
Furthermore, trust and anticipation outweigh fear, joy and sadness.
Regular readers of Sherlock Holmes novels should recognise the basic
elements that are inherent for the work of Arthur Conan Doyle.

\hypertarget{analysis---3-intertemporal-use-of-sentiments-conditional-mean}{%
\subsection{Analysis - 3: Intertemporal Use of Sentiments (Conditional
Mean)}\label{analysis---3-intertemporal-use-of-sentiments-conditional-mean}}

While the first two analytical steps related to the sentiments and
emotional states of words, the third step is about the conditional means
of their use over time - as known as the intertemporal use of them.
Therefore, the \emph{count(\ldots)} command needs to be adjusted in
order to count every \emph{sentiment} used in a \emph{line}. ´

\begin{Shaded}
\begin{Highlighting}[]
\NormalTok{nrc\_word\_counts }\OtherTok{\textless{}{-}}\NormalTok{ text\_tidy }\SpecialCharTok{\%\textgreater{}\%}
  \FunctionTok{inner\_join}\NormalTok{(}\FunctionTok{get\_sentiments}\NormalTok{(}\StringTok{"nrc"}\NormalTok{)) }\SpecialCharTok{\%\textgreater{}\%}
  \FunctionTok{count}\NormalTok{(line, sentiment, }\AttributeTok{sort =} \ConstantTok{TRUE}\NormalTok{) }\SpecialCharTok{\%\textgreater{}\%}
  \FunctionTok{ungroup}\NormalTok{()}
\end{Highlighting}
\end{Shaded}

\begin{verbatim}
## Joining, by = "word"
\end{verbatim}

Subsequently, object \emph{nrc\_word\_counts} is a tibble, indicating
not only the sentiments and emotional states within a \emph{line} but
also the total number \emph{n} of them and how often they are used
within that line. For example, line \emph{33} contains the sentiments
negative and the emotional states fear and sadness. In total, three
sentiments and emotional states are used in that line, but each of them
appears only once. As a result, the mean value for line \emph{33} would
be \emph{1}. Each line can be inspected by using the
\emph{subset(\ldots)} command:

\newpage

\begin{Shaded}
\begin{Highlighting}[]
\FunctionTok{subset}\NormalTok{(nrc\_word\_counts, line}\SpecialCharTok{==}\DecValTok{33}\NormalTok{)}
\end{Highlighting}
\end{Shaded}

\begin{verbatim}
## # A tibble: 3 x 3
##    line sentiment     n
##   <int> <chr>     <int>
## 1    33 fear          1
## 2    33 negative      1
## 3    33 sadness       1
\end{verbatim}

In order to highlight the dynamics regarding the appearance of
sentiments and emotional states over time, a smoothed slope will be
plotted. Smoothed slopes do not represent the actual values, like mean
values, but they do represent estimated values, in this case conditional
means, to indicate the development. Thus, primarily the course of the
slope should be interpreted and not individual data points. For example,
line \emph{133} contains a positive sentiment and three emotional states
- trust, anticipation and joy - whereas two of them appear eight times.
This paragraph (here: line) is about a narcissistic character, who is
talking about himself and people he would like to meet. Hence, the high
number of positive sentiments and emotional states. As a result, the
mean value for line \emph{133} should be much higher in respect to the
mean value of line \emph{33}, but the plotted slope will not reach out
to that specific mean value, it would only indicate a higher point in
the course of the slope. The \emph{subset(\ldots)} command highlights
line \emph{133} before plotting the slope:

\begin{Shaded}
\begin{Highlighting}[]
\FunctionTok{subset}\NormalTok{(nrc\_word\_counts, line}\SpecialCharTok{==}\DecValTok{133}\NormalTok{)}
\end{Highlighting}
\end{Shaded}

\begin{verbatim}
## # A tibble: 4 x 3
##    line sentiment        n
##   <int> <chr>        <int>
## 1   133 positive         8
## 2   133 trust            8
## 3   133 anticipation     2
## 4   133 joy              1
\end{verbatim}

The intertemporal use of sentiments and emotional states can be plotted
by using the \emph{ggplot(\ldots)} command, again. The \emph{span}
operator within the \emph{geom\_smooth(\ldots)} command specifies the
smoothness of the plotted slope and can be adjusted for a more detailed
representation:

\begin{Shaded}
\begin{Highlighting}[]
\FunctionTok{ggplot}\NormalTok{(}\AttributeTok{data =}\NormalTok{ nrc\_word\_counts, }\AttributeTok{mapping =} \FunctionTok{aes}\NormalTok{(}\AttributeTok{x =}\NormalTok{ line, }\AttributeTok{y =}\NormalTok{ n)) }\SpecialCharTok{+}
  \FunctionTok{geom\_smooth}\NormalTok{(}\AttributeTok{method=}\StringTok{"loess"}\NormalTok{, }\AttributeTok{formula=}\StringTok{"y\textasciitilde{}x"}\NormalTok{, }\AttributeTok{span=}\FloatTok{0.2}\NormalTok{) }\SpecialCharTok{+} 
  \FunctionTok{xlab}\NormalTok{(}\StringTok{"document (line)"}\NormalTok{) }\SpecialCharTok{+} 
  \FunctionTok{ylab}\NormalTok{(}\StringTok{"sentiment (conditional mean)"}\NormalTok{) }\SpecialCharTok{+}
  \FunctionTok{ggtitle}\NormalTok{(}\StringTok{"Intertemporal Use of Sentiments (Conditional Mean)"}\NormalTok{)}
\end{Highlighting}
\end{Shaded}

\includegraphics{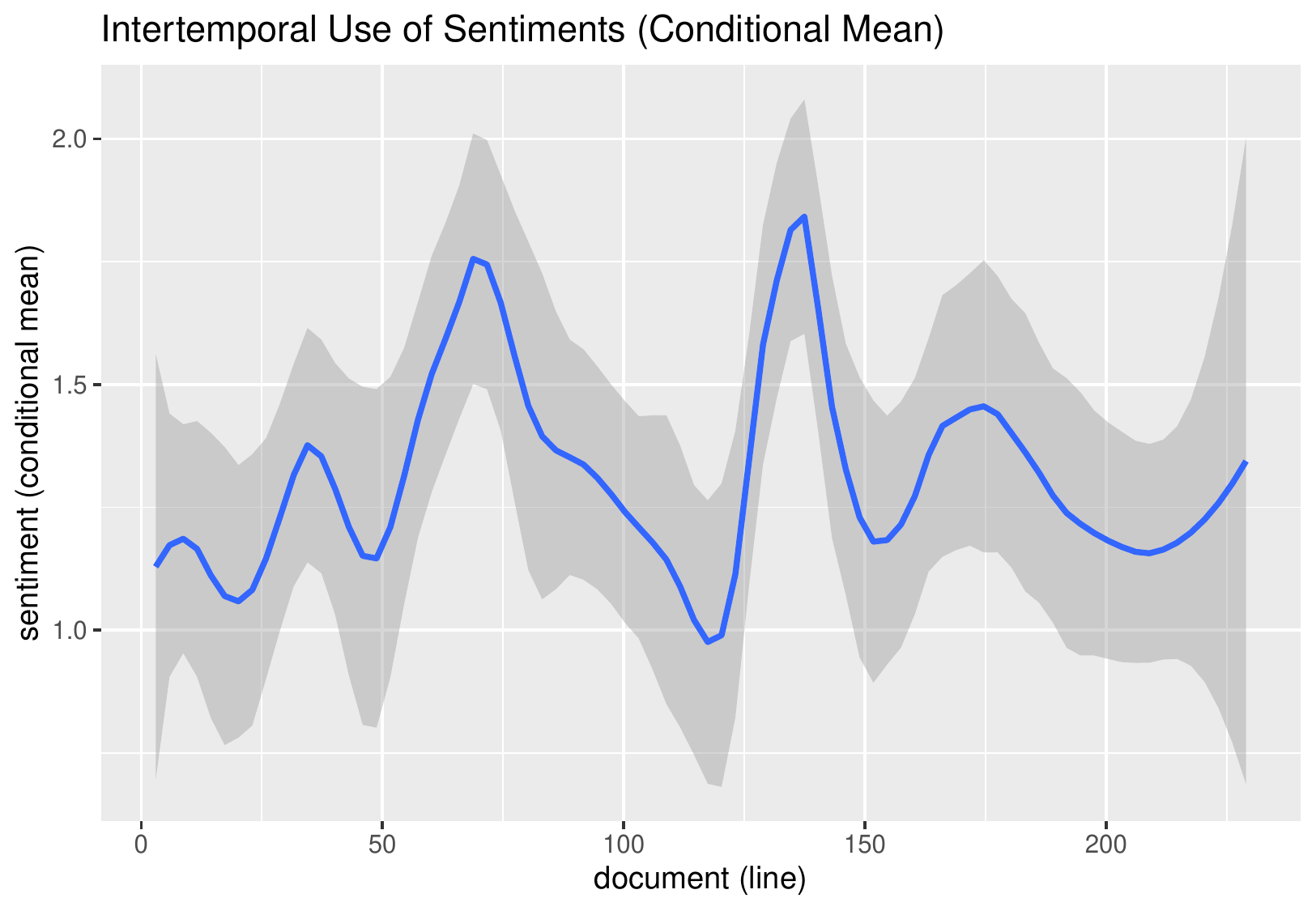}

The plot inidcates that fewer sentiments and emotional states (Y-axis)
are used at the beginning, the center and the end of this chapter than
in between (represented by lines on the X-axis). Other chapters of the
Sherlock Holmes novel could be analysed accordingly. However, this plot
just highlights the intertemporal use of sentiments and emotional states
as conditional means, regardless of whether they are positive or
negative. This aspect will be taken into account in the next step of
this tutorial.

\hypertarget{analysis---4-intertemporal-use-of-sentiments-score}{%
\subsection{Analysis - 4: Intertemporal Use of Sentiments
(Score)}\label{analysis---4-intertemporal-use-of-sentiments-score}}

Another way of analysing the intertemporal use of sentiments and
emotional states is based upon scores. A sentiment score results from
the sum of positive (1) and negative (-1) values in respect to the
underlying sentiment. Since the NRC Word-Emotion Association Lexicon
differentiates between positive and negative sentiments as well as eight
emotional states, a lexicon is required that differentiates only between
positive and negative sentiments: The Bing Sentiment Lexicon, written by
distinguished professor Bing Liu from the department of computer science
at the University of Illinois (Chicago), is suited for that task. By
specifying his lexicon within the \emph{inner\_join} command, a new
object called \emph{bing\_word\_counts} can be generated and inspected
by using the \emph{head(\ldots)} command:

\begin{Shaded}
\begin{Highlighting}[]
\NormalTok{bing\_word\_counts }\OtherTok{\textless{}{-}} \FunctionTok{bind\_rows}\NormalTok{(}
\NormalTok{  text\_tidy }\SpecialCharTok{\%\textgreater{}\%} 
    \FunctionTok{inner\_join}\NormalTok{(}\FunctionTok{get\_sentiments}\NormalTok{(}\StringTok{"bing"}\NormalTok{)) }\SpecialCharTok{\%\textgreater{}\%}
    \FunctionTok{mutate}\NormalTok{(}\AttributeTok{method =} \StringTok{"Bing et al."}\NormalTok{))}
\end{Highlighting}
\end{Shaded}

\begin{verbatim}
## Joining, by = "word"
\end{verbatim}

\newpage

\begin{Shaded}
\begin{Highlighting}[]
\FunctionTok{head}\NormalTok{(bing\_word\_counts)}
\end{Highlighting}
\end{Shaded}

\begin{verbatim}
## # A tibble: 6 x 4
##    line word      sentiment method     
##   <int> <chr>     <chr>     <chr>      
## 1     4 love      positive  Bing et al.
## 2     5 cold      negative  Bing et al.
## 3     5 precise   positive  Bing et al.
## 4     5 admirably positive  Bing et al.
## 5     6 balanced  positive  Bing et al.
## 6     6 perfect   positive  Bing et al.
\end{verbatim}

Within this object a sentiment is designated to each word and the
\emph{ifelse(\ldots)} command assigns values of \emph{1} and \emph{-1}
accordingly. A few commands are required in order to generate a data
frame that aggregates these values for each line. Therefore, after the
\emph{cbind(\ldots)} command has focused the relevant variables but
eliminated their labels, the lines will be labeled \emph{var1} and the
sum of the values \emph{var2}. After transformation into a data frame by
using the \emph{as.data.frame(\ldots)} command the
\emph{aggregate(\ldots)} command generates the necessary object
\emph{sentiment\_sum\_df}, which can be sorted in respect to the lines
within \emph{var1}:

\begin{Shaded}
\begin{Highlighting}[]
\NormalTok{sentiment\_sum }\OtherTok{\textless{}{-}} \FunctionTok{ifelse}\NormalTok{(bing\_word\_counts}\SpecialCharTok{$}\NormalTok{sentiment }\SpecialCharTok{==} \StringTok{"positive"}\NormalTok{, }\DecValTok{1}\NormalTok{, }\SpecialCharTok{{-}}\DecValTok{1}\NormalTok{)}
\NormalTok{sentiment\_sum\_df }\OtherTok{\textless{}{-}} \FunctionTok{cbind}\NormalTok{(bing\_word\_counts}\SpecialCharTok{$}\NormalTok{line, sentiment\_sum)}
\FunctionTok{colnames}\NormalTok{(sentiment\_sum\_df) }\OtherTok{\textless{}{-}} \FunctionTok{c}\NormalTok{(}\StringTok{\textquotesingle{}var1\textquotesingle{}}\NormalTok{, }\StringTok{\textquotesingle{}var2\textquotesingle{}}\NormalTok{)}
\NormalTok{sentiment\_sum\_df }\OtherTok{\textless{}{-}} \FunctionTok{as.data.frame}\NormalTok{(sentiment\_sum\_df)}
\NormalTok{sentiment\_sum\_df }\OtherTok{\textless{}{-}} \FunctionTok{aggregate}\NormalTok{(sentiment\_sum\_df}\SpecialCharTok{$}\NormalTok{var2,}
  \AttributeTok{by=}\FunctionTok{list}\NormalTok{(}\AttributeTok{line=}\NormalTok{sentiment\_sum\_df}\SpecialCharTok{$}\NormalTok{var1), }\AttributeTok{FUN=}\NormalTok{sum)}
\end{Highlighting}
\end{Shaded}

The \emph{head(\ldots)} command provides insights into the data frame
called \emph{sentiment\_sum\_df}. For each line with sentiments the
score of these sentiments (x) is assigned as sum of all positive (1) and
negative (-1) sentiments:

\begin{Shaded}
\begin{Highlighting}[]
\FunctionTok{head}\NormalTok{(sentiment\_sum\_df)}
\end{Highlighting}
\end{Shaded}

\begin{verbatim}
##   line  x
## 1    4  1
## 2    5  1
## 3    6  2
## 4    7  1
## 5    8  0
## 6    9 -1
\end{verbatim}

The related plot is called via the \emph{ggplot(\ldots)} command as in
the previous steps of this tutorial. Again, the X-axis relates to the
lines in the text document. This time, the Y-axis relates to the
previously generated sentiment score. The result is an intertemporal
representation of these scores line by line:

\begin{Shaded}
\begin{Highlighting}[]
\FunctionTok{ggplot}\NormalTok{(}\AttributeTok{data =}\NormalTok{ sentiment\_sum\_df, }\AttributeTok{mapping =} \FunctionTok{aes}\NormalTok{(}\AttributeTok{x =}\NormalTok{ line, }\AttributeTok{y =}\NormalTok{ x)) }\SpecialCharTok{+} 
  \FunctionTok{geom\_smooth}\NormalTok{(}\AttributeTok{method=}\StringTok{"loess"}\NormalTok{, }\AttributeTok{formula=}\StringTok{"y\textasciitilde{}x"}\NormalTok{, }\AttributeTok{span=}\FloatTok{0.2}\NormalTok{) }\SpecialCharTok{+} 
  \FunctionTok{xlab}\NormalTok{(}\StringTok{"document (line)"}\NormalTok{) }\SpecialCharTok{+} 
  \FunctionTok{ylab}\NormalTok{(}\StringTok{"sentiment (score)"}\NormalTok{) }\SpecialCharTok{+}
  \FunctionTok{ggtitle}\NormalTok{(}\StringTok{"Intertemporal Use of Sentiments (Score)"}\NormalTok{)}
\end{Highlighting}
\end{Shaded}

\includegraphics{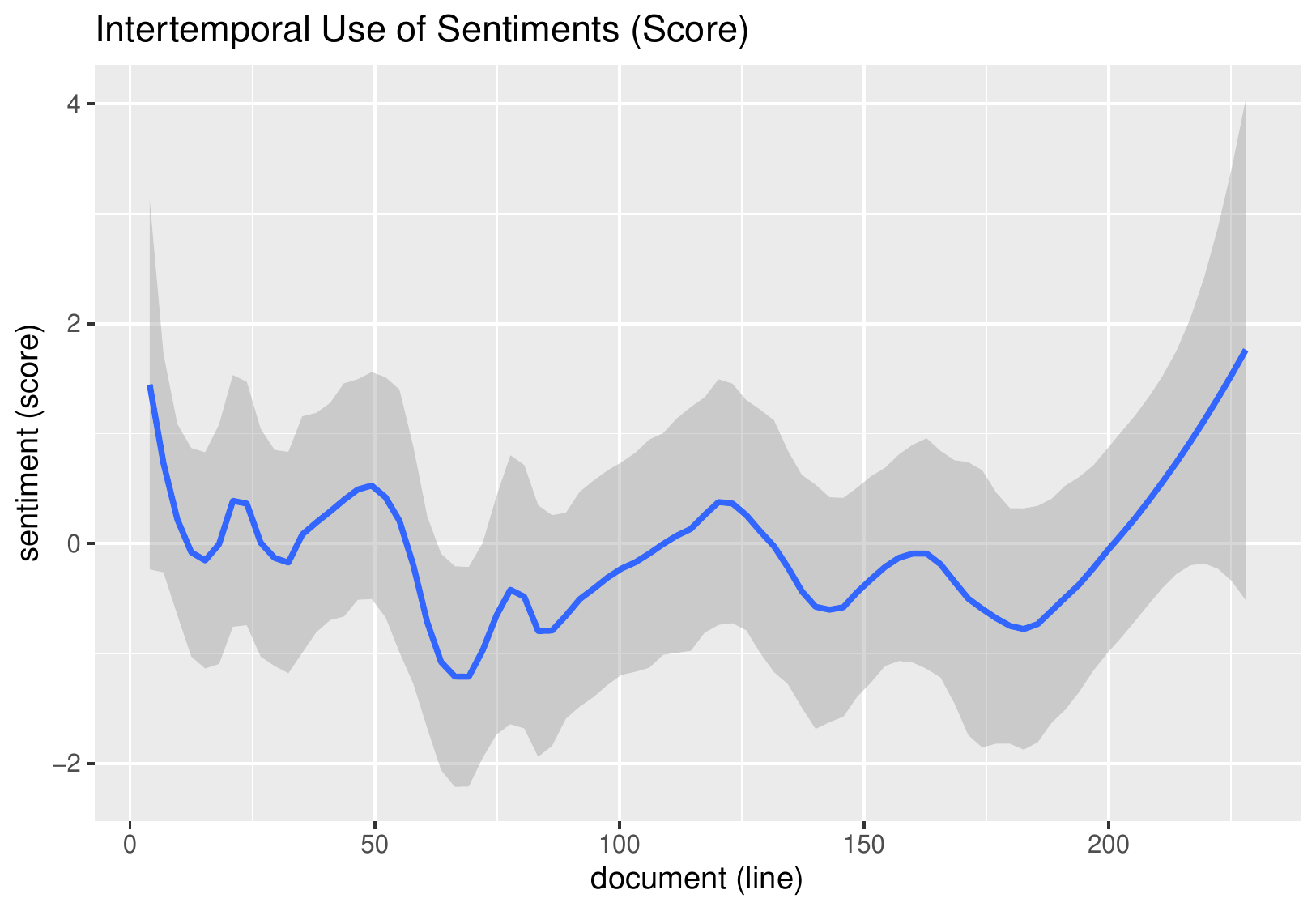}

As a result, this Sherlock Holmes chapter seems to have positive
sentiments at the beginning and an increasing sentiment at the end,
while there appears to be a dramaturgical downscaling during the chapter
to highlight the tensions in solving the crime.

\hypertarget{application-example-comparing-political-speeches}{%
\subsection{Application Example: Comparing Political
Speeches}\label{application-example-comparing-political-speeches}}

The last chapter of this tutorial is about comparing two political
speeches. For that purpose, English-language versions of two
simultaneous speeches on a specific issue relating the war in Ukraine
were selected. One speech is from Jens Stoltenberg (Secretary General of
NATO) and the other speech - that is supposed to be an answer to the
first one - is from Sergey Viktorovich Lavrov (Minister of Foreign
Affairs, Russian Federation). Basic and advanced sentiment analyses
enable a comparison of the linguistic composition of both speeches.

All commands can be used as in the previous steps of this tutorial. In
order to systematically summarise the graphical findings, a dashboard
with all four plots at once is generated by using the previously
installed package \emph{gridExtra} and the \emph{grid.arrange(\ldots)}
command. This requires that all previous plots be created as new
objects, for example: \emph{plot1 \textless- ggplot(\ldots)}. The
assignment arrow should always be on the first line of the commands that
contain the \emph{ggplot(\ldots)} command and each new object should be
the first word in that command. This makes it possible to display all
plots from \emph{plot1} to \emph{plot4} in one dashboard:

\begin{Shaded}
\begin{Highlighting}[]
\FunctionTok{grid.arrange}\NormalTok{(plot1, plot2, plot3, plot4)}
\end{Highlighting}
\end{Shaded}

The corresponding output for both speeches is plotted on the next page:

\newpage

The dashboard regarding the speech of Jens Stoltenberg:

\begin{center}\includegraphics[width=0.9\linewidth]{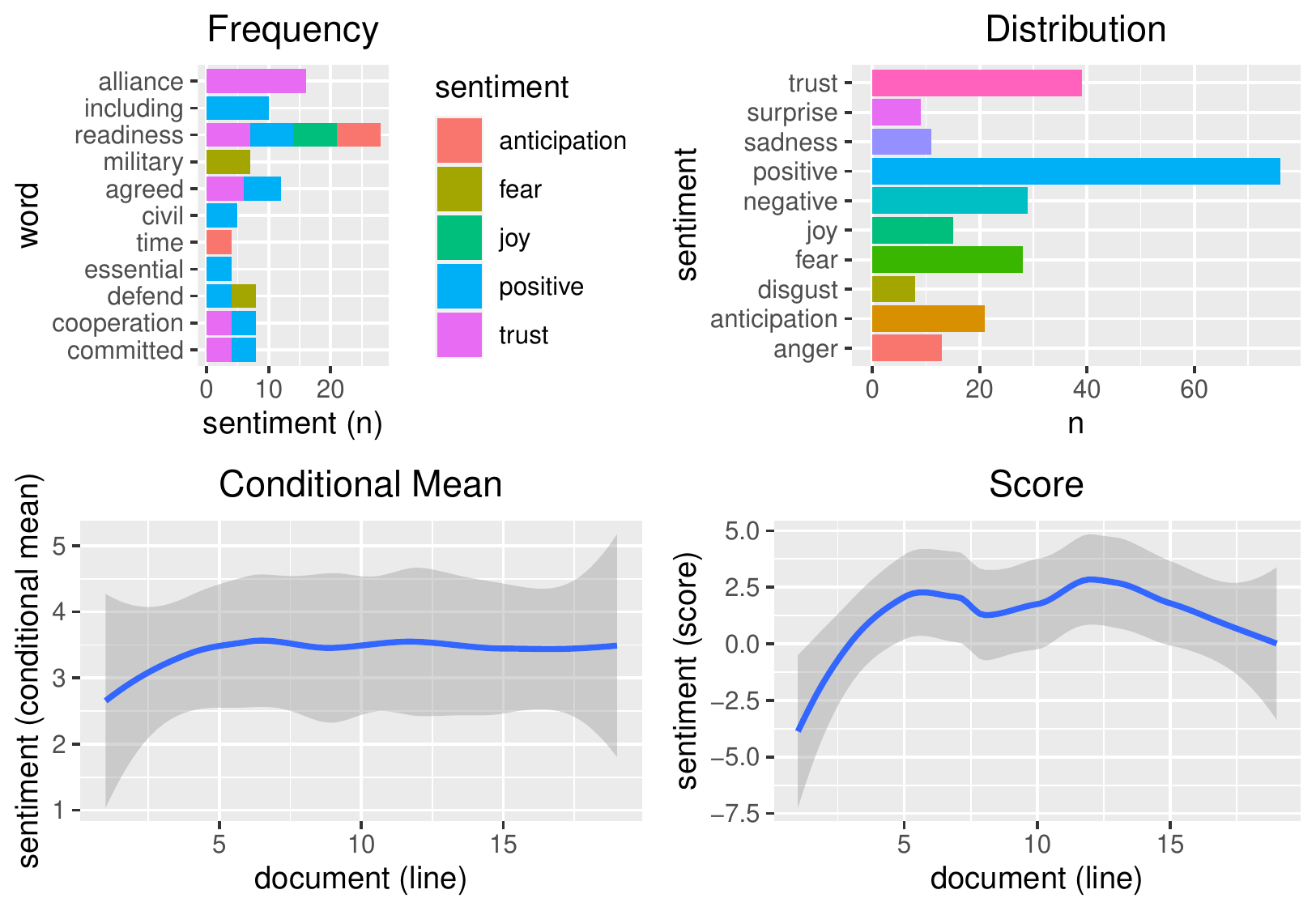} \end{center}

In comparison, the dashboard regarding the speech of Sergey Viktorovich
Lavrov:

\begin{center}\includegraphics[width=0.9\linewidth]{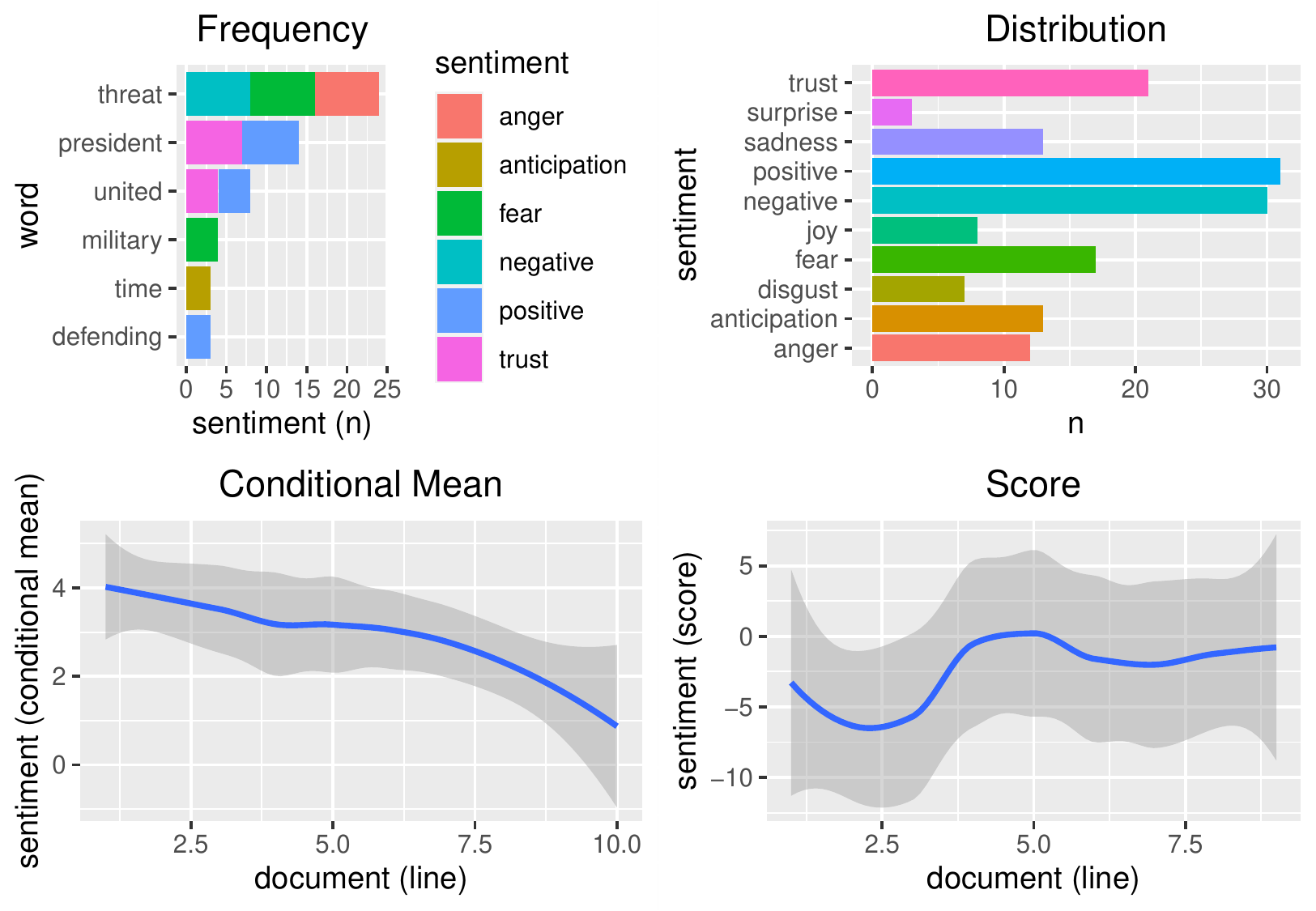} \end{center}

\newpage

The plots above indicate that Jens Stoltenberg focuses the readiness and
the alliance of NATO states by expressing trust and anticipation as
emotional states (Frequency). There is also a strong use of positive
sentiments within his speech (Distribution). Overall, his speech is
characterised by an estimated use of three sentiments and emotional
states per paragraph, while his speech begins with an estimate of only
two sentiments and emotional states (Conditional Mean). The beginning of
his speech can be characterised as negative, which changes to a positive
sentiment score as the speech progresses (Score). Finally, his speech
ends with a neutral sentiment score.

Sergey Viktorovich Lavrov, on the other hand, refers to a threat for the
Russian Federation and his trust in the president of the Russian
Federation (Frequency). The threat is expressed with the emotional
states fear and anger, thus a negative sentiment is inherent. When it
comes to the distribution of sentiments and emotional states, positive
and negative sentiments seem to be equally used, while trust and
anticipation counterbalance the emotional states fear and anger
(Distribution). His speech begins with an estimate of four sentiments
and emotional states but he seems to use fewer sentiments and emotional
states at the end of his speech (Conditional Mean). In total, his
sentiment score remains negative with a slight increase at the center of
his speech (Score).

Based on such findings, it seems necessary to inspect the corresponding
paragraphs (here: lines) which include the highlighted words, their
sentiments and emotional states, but also the intertemporal use of them.
For example, each turning point and saddle point of the slopes could be
inspected line by line to validate these findings. Finally, this is
primarily a technique in order to speed up the process of analysing text
documents and qualitative data, rather than a possibility of reaching to
final conclusions.

\hypertarget{sources}{%
\subsection{Sources}\label{sources}}

\begin{itemize}
\item
  Auguie, Baptise \& Anton Antonov (2017): ``gridExtra: Miscellaneous
  Functions for''Grid'' Graphics''. Online:
  \url{https://cran.r-project.org/package=gridExtra}
\item
  Bache, Stefan Milton \& Hadley Wickham (2020): ``magrittr: A
  Forward-Pipe Operator for R''. Online:
  \url{https://cran.r-project.org/package=magrittr}
\item
  De Queiroz, Gabriela; Fay, Colin; Hvitfeldt, Emil; Keyes, Os; Misra,
  Kanishka; Mastny, Tim; Erickson, Jeff; Robinson, David \& Julia Silge
  (2022): ``tidytext: Text Mining using''dplyr'', ``ggplot2'', and Other
  Tidy Tools''. Online:
  \url{https://cran.r-project.org/package=tidytext}
\item
  Hamborg, Felix \& Karsten Donnay (2021): ``NewsMTSC: A Dataset for
  (Multi-)Target-dependent Sentiment Classification in Political News
  Articles''. Proceedings of the 16th Conference of the European Chapter
  of the Association for Computational Linguistics.
\item
  Klinkhammer, Dennis (2020): ``Analysing Social Media Network Data with
  R: Semi-Automated Screening of Users, Comments and Communication
  Patterns''. Online: \url{https://arxiv.org/abs/2011.13327}
\item
  Korkontzelos, Ioannis; Nikfarjam, Azadeh; Shardlow, Matthew; Sarker,
  Abeed; Ananiadou, Sophia \& Graciela Gonzalez (2016): ``Analysis of
  the effect of sentiment analysis on extracting adverse drug reactions
  from tweets and forum posts''. Journal of Biomedical Informatics (62):
  148--158.
\item
  Mohammad, Saif M. (2020): ``Sentiment Analysis: Automatically
  Detecting Valence, Emotions, and Other Affectual States from Text''.
  Online: \url{https://arxiv.org/abs/2005.11882}
\item
  Ogneva, Maria (2010): ``How Companies Can Use Sentiment Analysis to
  Improve Their Business''. Online:
  \url{https://mashable.com/archive/sentiment-analysis}
\item
  Tanoli, Irfan; Pais, Sebastiao; Cordeiro, Joao \& Muhammad Luqman
  Jamil (2022): ``Detection of Radicalisation and Extremism Online: A
  Survey''. Online:
  \url{https://assets.researchsquare.com/files/rs-1185415/v1_covered.pdf}
\item
  Tumasjan, Andranik; Sprenger, Timm O.; Sandner, Philipp G. \& Isabelle
  M. Welpe (2010): ``Predicting Elections with Twitter: What 140
  Characters Reveal about Political Sentiment''. Proceedings of the
  Fourth International AAAI Conference on Weblogs and Social Media.
\item
  Wickham, Hadley (2019): ``stringr: Simple, Consistent Wrappers for
  Common String Operations''. Online:
  \url{https://cran.r-project.org/package=stringr}
\item
  Wickham, Hadley; François, Roman \& Kirill Müller (2022): ``dplyr: A
  Grammmar of Data Manipulation''. Online:
  \url{https://cran.r-project.org/package=dplyr}
\item
  Wickham, Hadley; Chang, Winston; Henry, Lionel; Lin Pedersen, Thomas;
  Takahashi, Kohske; Wilke, Claus; Woo, Kara; Yutani, Hiroaki \& Dewey
  Dunnington (2022): ``ggplot2: Create Elegant Data Visualisations Using
  the Grammar of Graphics''. Online:
  \url{https://cran.r-project.org/package=ggplot2}
\item
  Wood, Ian B.; Varela, Pedro Leal; Bollen, Johan; Rocha, Luis M. \&
  Joana Gonçalves-Sá (2017): ``Human Sexual Cycles are Driven by Culture
  and Match Collective Moods''. Online:
  \url{https://arxiv.org/abs/1707.03959}
\end{itemize}

\hypertarget{author-affiliations-and-materials-on-github}{%
\subsection{Author, Affiliations and Materials on
GitHub}\label{author-affiliations-and-materials-on-github}}

Dennis Klinkhammer is Professor for Empirical Research at the FOM
University of Applied Sciences. He advises public as well as
governmental organisations on the application of multivariate statistics
and limitations of artificial intelligence by providing introductions to
Python and R: \url{https://www.statistical-thinking.de}

All codes required for the basic and advanced Sentiment Analysis with R
can be accessed on GitHub:
\url{https://github.com/statistical-thinking/sentiment-analysis}

\end{document}